\documentclass[sigconf]{acmart}
\usepackage{booktabs}
\usepackage{multirow}
\usepackage{makecell}
\usepackage{subfigure}
\usepackage{amsthm}
\usepackage{enumitem}
\usepackage{amsmath, amsthm}
\usepackage{color}
\usepackage{bbm}

\usepackage{algorithm}
\usepackage{algpseudocode}

\usepackage[normalem]{ulem}
\useunder{\uline}{\ul}{}
\theoremstyle{definition}

\newcommand{\ie}{\textit{i}.\textit{e}.}

\AtBeginDocument{%
  }

\copyrightyear{2025}
\acmYear{2025}
\setcopyright{acmlicensed}
\acmConference[MM '25] {Proceedings of the 33rd ACM International Conference on Multimedia}{October 27--31, 2025}{Dublin, Ireland.}
\acmBooktitle{Proceedings of the 33rd ACM International Conference on Multimedia (MM '25), October 27--31, 2025, Dublin, Ireland}
\acmISBN{979-8-4007-2035-2/2025/10}
\acmDOI{10.1145/3746027.3755245}

\begin{document}

\title{Can LLMs Find Fraudsters? Multi-level LLM Enhanced Graph Fraud Detection}

\author{Tairan Huang}
\email{tairanhuang@csu.edu.cn}
\affiliation{%
  \institution{Central South University}
  \city{Changsha}
  \country{China}}

\author{Yili Wang}
\email{yiliwang@hkust-gz.edu.cn}
\affiliation{%
  \institution{Hongkong University of Science and Technology (Guangzhou)}
  \city{Guangzhou}
  \country{China}}

\author{Qiutong Li}
\email{qiutonglee@csu.edu.cn}
\affiliation{%
  \institution{Central South University}
  \city{Changsha}
  \country{China}}
  
\author{Changlong He}
\email{frankemail@csu.edu.cn}
\affiliation{%
  \institution{Central South University}
  \city{Changsha}
  \country{China}}

\author{Jianliang Gao}
\authornote{Corresponding author.}
\email{gaojianliang@csu.edu.cn}
\affiliation{%
  \institution{Central South University}
  \city{Changsha}
  \country{China}}

\renewcommand{\shortauthors}{Tairan Huang, Yili Wang, Qiutong Li, Changlong He, and Jianliang Gao}

\begin{abstract}
 Graph fraud detection has garnered significant attention as Graph Neural Networks (GNNs) have proven effective in modeling complex relationships within multimodal data.
    However, existing graph fraud detection methods typically use preprocessed node embeddings and predefined graph structures to reveal fraudsters, which ignore the rich semantic cues contained in raw textual information. 
    Although Large Language Models (LLMs) exhibit powerful capabilities in processing textual information, it remains a significant challenge to perform multimodal fusion of processed textual embeddings with graph structures.
    In this paper, we propose a \textbf{M}ulti-level \textbf{L}LM \textbf{E}nhanced Graph Fraud \textbf{D}etection framework called MLED.
    In MLED, we utilize LLMs to extract external knowledge from textual information to enhance graph fraud detection methods. 
    To integrate LLMs with graph structure information and enhance the ability to distinguish fraudsters, we design a multi-level LLM enhanced framework including type-level enhancer and relation-level enhancer. 
    One is to enhance the difference between the fraudsters and the benign entities, the other is to enhance the importance of the fraudsters in different relations.
    The experiments on four real-world datasets show that MLED achieves state-of-the-art performance in graph fraud detection as a generalized framework that can be applied to existing methods.
\end{abstract}




\maketitle
\section{Introduction}

Modern multimedia ecosystems are inherently manifested as complex multimodal graphs, such as social media platforms \cite{social}, e-commerce systems \cite{recommendation}, and video knowledge graphs \cite{knowledge}, where heterogeneous nodes (users, products, multimedia content) interact through both structural connections and rich multimodal signals. 
The emerging paradigm of multimodal graph neural networks has demonstrated unprecedented capabilities in modeling these intricate relationships, particularly when augmented with Large Language Models (LLMs) that provide cross-modal semantic grounding for visual \cite{visual}, textual \cite{textual}, and behavioral data \cite{behavior}. 
However, the proliferation of sophisticated fraud patterns-ranging from fake multimedia reviews to AI-generated deceptive content-poses critical challenges to multimodal system integrity. 
These malicious campaigns often exploit semantic differences between graph topologies and unstructured modalities to undermine the reliability of multimedia recommendation and content understanding systems.
Therefore, the development of robust multimodal graph fraud detection mechanisms is essential to ensure the veracity of cross-modal correlations while maintaining the complementary advantages of structural and semantic representations.

\begin{figure}[th]
	\centering
    \includegraphics[width=\linewidth]{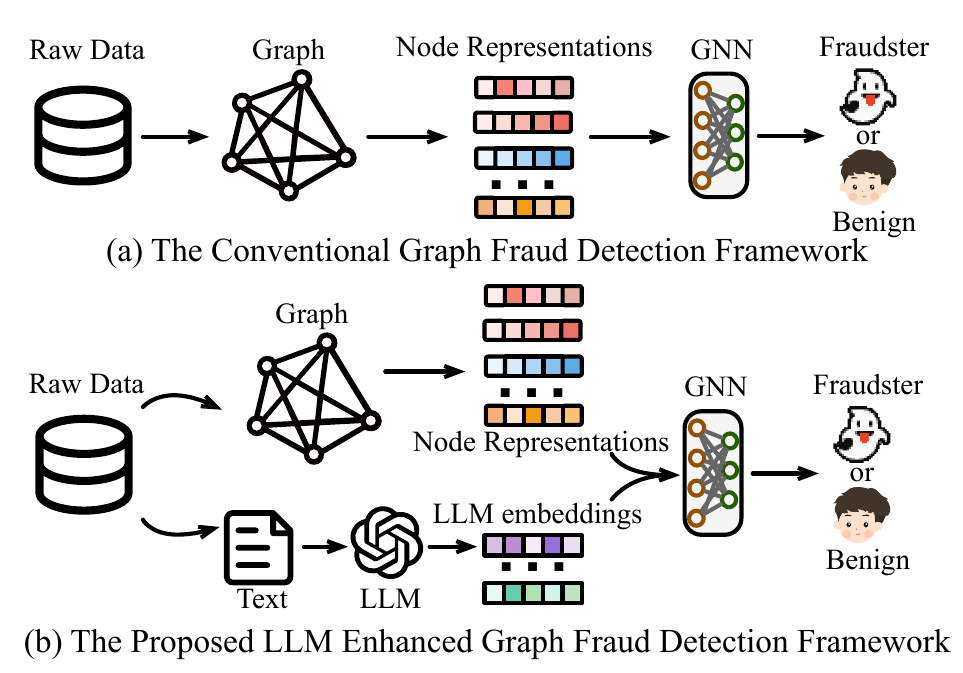}
	\caption{Workflow comparison. The conventional graph fraud detection methods only utilize GNNs for model training. In contrast, our framework combines textual and graph structural information through LLMs and GNNs to enhance the ability to distinguish fraudsters.}
	\label{motivation}
\end{figure}

Conventional graph-based fraud detection methods \cite{CARE-GNN, GAGA} reveal fraudsters by aggregating the features of their neighbors, which mostly rely on the assumption of homophily (\ie, connected nodes usually belong to the same class).
Recent studies have seen approaches that modify the graph structure to satisfy the homophily assumption \cite{homophily, GDN}, while others argue that extracting heterogeneous information from multiple relations is sufficiently beneficial \cite{GraphConsis, BWGNN, GHRN}.
These graph fraud detection methods reveal fraudsters by different designs of GNNs as shown in Figure 1(a).
Besides, with the emergence of powerful LLMs such as \emph{GPT} and \emph{Llama}, these models demonstrate the ability to reason and leverage general knowledge.
An ideal solution is to combine textual and graph structural information in graph fraud detection through LLMs.

However, there are two major challenges for graph fraud detection with LLMs. 
$\textbf{1)}$ \textbf{How to integrate LLMs with graph structural information for fraud detection.}
Recently, several studies have attempted to use powerful LLMs to process the raw textual information of nodes, which accomplishes downstream tasks in graph-based applications.
The node textual information processed by LLMs can effectively improve the performance of GNN-based methods when combined with the graph structure.
However, the computational and memory costs to process the textual information individually for each node with LLMs can become prohibitive in large node-dense graphs.
Moreover, in graph fraud detection tasks, obtaining the raw textual information of the nodes is a significant challenge due to issues such as user privacy.
Therefore, how to efficiently combine LLMs with graph structures for fraud detection is an urgent problem.
$\textbf{2)}$ \textbf{How to enhance the ability to distinguish fraudsters with LLMs.}
Considering that fraudsters are typically connected to benign entities in the graph, the messages from fraud neighbors can easily be obscured during message propagation.
The previous graph fraud detection methods design various GNN architectures to enhance the distinctions between fraudsters and benign entities.
However, these types of methods can only enhance the ability to distinguish fraudsters in limited ways.
Therefore, it is a promising way to enrich the representation of nodes with textual information to better distinguish between fraudsters and benign entities.

To tackle these two challenges, we propose a \textbf{M}ulti-level \textbf{L}LM \textbf{E}nhanced Graph Fraud \textbf{D}etection framework named MLED.
In MLED, we utilize LLMs to extract external knowledge from textual information to enhance graph fraud detection methods, as illustrated in Figure 1(b).
We design a multi-level LLM enhanced framework including type-level enhancer and relation-level enhancer. 
In the type-level enhancer, we calculate node-type embeddings with LLMs to enhance the distinction between fraudsters and benign entities.
Meanwhile, we equally adopt LLMs to calculate relation embeddings to enhance the importance of the fraudsters in different relations by the relation-level enhancer.
The experiments on two real-world datasets demonstrate that our framework, as a general framework applicable to existing methods, significantly improves the performance of state-of-the-art (SOTA) methods on graph fraud detection tasks.
Our contributions are summarized as follows:
\begin{itemize}[leftmargin=*]
	\item \textbf{First exploration of fraud detection with LLMs in graph.}
    To the best of our knowledge, we are the first to use the powerful LLMs for graph fraud detection tasks.
    Our innovation is the multimodal fusion of textual information with graph structures to inspire fraud detection, which improves model performance and introduces only a small time overhead.
	\item \textbf{Novel multi-level LLM enhanced framework for graph fraud detection.} 
    We pioneer to present the multi-level LLM enhanced framework, MLED, including two different level enhancers. One is to enhance the difference between the fraudsters and the benign entities, the other is to enhance the importance of the fraudsters in different relations.
	\item\textbf{SOTA performance.} We conduct extensive experiments on four real-world datasets to demonstrate the effectiveness of MLED, achieving state-of-the-art performance on graph fraud detection tasks.
\end{itemize}


\begin{figure*}[th]
	\centering
	\includegraphics[width=\linewidth]{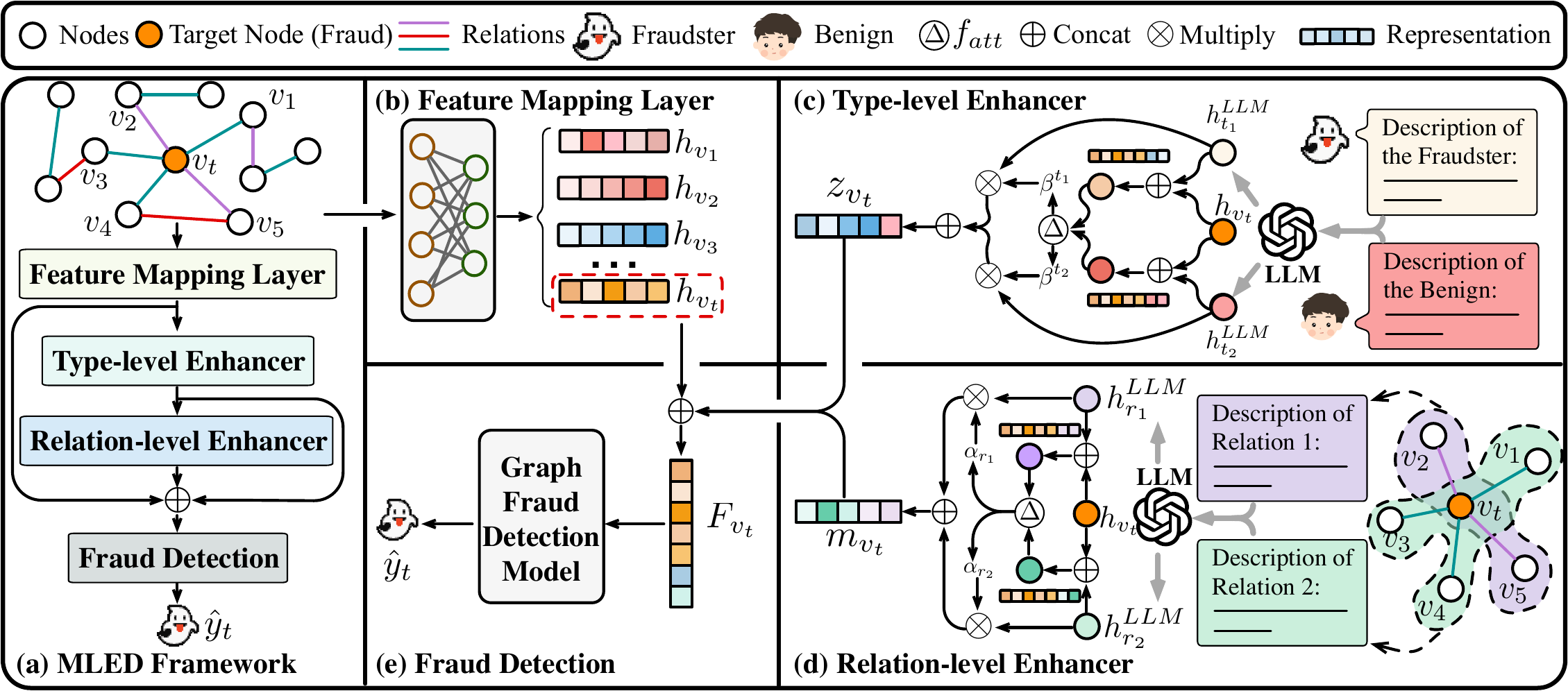}
	\caption{The overall framework of MLED. 
    }
	\label{framework}
\end{figure*}

\section{Related Works}
\subsection{Graph Fraud Detection}
Existing graph fraud detection methods can be systematically categorized into four paradigms based on their technical foundations.
The traditional methods leverage the homophily principle, assuming that benign nodes maintain topological and attribute consistency with their neighbors \cite{GAD_1, GAD_2, GAD_3, GAD_4}.
CARE-GNN \cite{CARE-GNN} employs reinforcement learning to prune heterophilic edges, while GraphConsis \cite{GraphConsis} models attribute-structure consistency through contrastive learning. 
However, these methods struggle with camouflage attacks, in which fraudsters deliberately imitate benign behavioral patterns.
Methods that leverage spectral graph theory and deep representation fusion have shown promise in capturing global fraud patterns \cite{spe_1, spe_2, spe_3}.
BWGNN \cite{BWGNN} and GHRN \cite{GHRN} detect fraudsters through graph signal processing, which utilizes spectral clustering in adjacency matrices.
SaVAE-SR \cite{spectralres} and SplitGNN \cite{split} analyze the spectral distribution across varying degrees of heterogeneity, which use edge classifiers to segment the original graph and a flexible bandpass graph filter to learn representations.
Although effective in detecting structural fraudsters, these methods often fail to capture semantic inconsistencies between node attributes and topological positions.
Some researchers have focused on improving fraud detection in heterogeneous graphs, such as MVHGN \cite{MVHGNN} and CATS \cite{Cross} to align cross-platform user embeddings using graph similarity learning.
Furthermore, fraudsters typically evolve their strategies over time, which has motivated research on dynamic graphs.
SAD \cite{SAD} and SLADE \cite{SLADE} propose dynamic graph fraud detection frameworks to discover potential fraudsters in evolving graph streams.
However, these methods primarily operate in modality-isolated regimes, focusing on either structural patterns or single-modality attributes. 
The fundamental challenge of integrating discrete graph structures with continuous multimodal semantics remains underexplored, particularly in constructing cross-modal dependency graphs and resolving semantic-topological conflicts.

\subsection{Large Language Models on Graphs}
The integration of large language models with graph neural networks has become a vibrant research frontier.
The graph architectural augmentation methods integrate LLMs directly into the GNN framework to enhance semantic reasoning \cite{llm_1, llm_2}.
GraphBERT \cite{Graphbert} pioneers this direction by replacing traditional GNN layers with transformer modules pre-trained on textual node attributes. 
GIANT \cite{GIANT} fine-tunes language models using a novel self-supervised learning framework with XR-Transformers to tackle extreme multi-label categorization in link prediction.
WalkLM \cite{llm_2} and TouchUp-G \cite{touchup} adopt a similar approach, which uses LLMs for link prediction to enhance their perception of structural information.
However, these methods require significant overhead when dealing with large-scale datasets, although they are able to enhance the semantic reasoning ability of GNNs.
Several methods transfer LLM-derived knowledge to GNNs through offline knowledge distillation. 
TextGNN \cite{textgnn} aligns GNN embeddings with LLM-generated text prototypes, while KDDG \cite{kddg} distills LLM reasoning paths into graph topology refinement rules.
These methods decouple the linguistic and structural learning phases with high computational efficiency but fail to capture real-time interactions between textual semantics and graph evolution, which is a key limitation in detecting multimodal fraud patterns resulting from coordinated attribute-structure attacks.
The prompt-based methods bridge LLMs and GNNs through task-specific instructions.
GraphPrompt \cite{graphprompt} and GPPT \cite{gppt} design structural prompts to guide LLMs in generating node-aware textual representations.
To address the interpretability challenges of GNNs, Pan et al. \cite{pan} construct an interpreter model that integrates LLMs and GNNs to enhance inference and generalization capabilities in textual attribute graph tasks.
LLMExplainer \cite{LLM_ex} embeds LLMs as Bayesian inference modules to evaluate and refine explanation subgraphs.
This framework mitigates learning biases caused by sparse annotations, which integrates LLM-generated confidence scores into variational inference. 
However, current LLM-GNN integrations primarily operate in passive fusion modes. 
In real-time reasoning, existing methods lack mechanisms to dynamically associate linguistic fraudster (e.g., fraud descriptions) with structural suspicions.

\section{Methodology}
In this section, we introduce the detailed design of MLED. 
The overall framework is illustrated in Figure \ref{framework}.
We first conduct the preliminaries to demonstrate notations and problem definition in graph fraud detection, and then present the modules of the proposed MLED framework.

\subsection{Preliminaries}
The multi-relation graph is defined as $\mathcal{G = \{V, \{E}_r|^R_{r=1}\}\}$, where $\mathcal{V} = \{v_1, v_2, \ldots, v_N\}$ is the set of nodes, $R $ denotes the number of relations, and $e_{i, j}^r \in \mathcal{E}_r$ indicates an edge between nodes $v_i$ and $v_j$ under relation $r$. 
For all $N$ nodes, the set of node features is defined as $\mathcal{X}  = \{x_1, x_2, \ldots, x_N\} \in \mathbb{R}^{N \times D}$, where each node $v_t$ has a $D$-dimensional feature vector. 
$Y = \{y_1, y_2, \ldots, y_N\}$ denotes the labels of nodes, where $y_n (0/1)$ is the ground-truth label, with 0 indicating a benign user and 1 indicating a fraudster.
Then, we formalize the graph-based fraud detection problem as follows:
\begin{definition}
	\emph{(Graph Fraud Detection)}. 
	Given the multi-relation graph $\mathcal{G}$, node features $\mathcal{X}$ and labels $Y$, the objective is to determine whether a given target node $v_t$ is suspicious. 
    In fraud detection tasks, the number of fraudulent nodes is often significantly smaller than that of benign ones, creating an imbalance in the dataset.
    This leads to the problem being formulated as an imbalanced binary fraud detection task.
    Our goal is to learn a function $\hat{y}_t = \Gamma (\mathcal{G}, \mathcal{X}, \hbar)$, where $\hat{y}_t$ represents the predicted classification of the target node $v_t$, and $\hbar$ refers to the model parameters associated with function $\Gamma$.
\end{definition}


\subsection{Feature Mapping Layer}
Previous work shows that the entanglement of filters and weight matrices may be harmful to the performance of the model \cite{mapping}. 
Inspired by this, the fully connected layer is leveraged to project input nodes into the same latent representation:
\begin{equation}
	h_{v_t} = \sigma(W_hx_t + b_h),
\end{equation} 
where $h_{v_t} \in \mathbb{R}^U$ indicates the embedding of node $v_t$, $\sigma$ is \emph{LeakyReLU}
, $W_h \in \mathbb{R}^{U\times D}$ is the trainable parameter and $b_h \in \mathbb{R}^U$ is the bias. 

\subsection{Type-level Enhancer}
We propose a novel type-level enhancer based on LLMs to enhance the difference between the fraudsters and the benign entities.
The prompt-based learning provides textual instructions for large language models, which allows LLMs to adapt to different tasks and various data domains.
 
\begin{figure}[th]
	\centering
    \includegraphics[width=\linewidth]{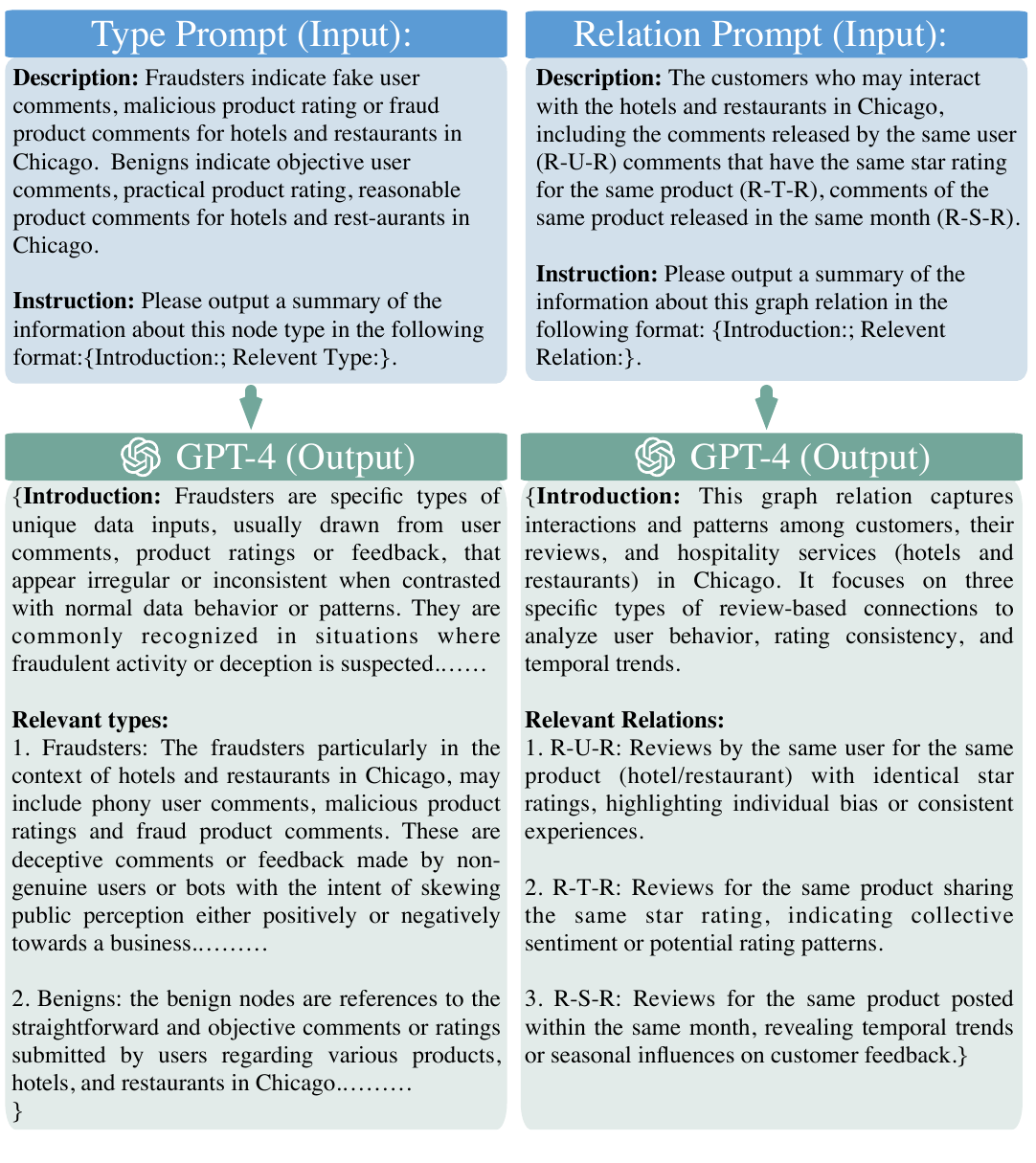}
	\caption{An example of constructing prompts for type-level enhancer and relation-level enhancer on YelpChi dataset.}
	\label{gpt}
\end{figure}

To integrate LLMs with graph structures for fraud detection, we first construct a brief text-based description of each node type and an instruction that requires output in a fixed format, using these as prompts for the types: $P(t_n)=$ \{Introduction to the type $t_n$; Instruction\}.
The example of prompt construction can be found in Figure \ref{gpt}.
Then, LLM (\emph{GPT-4}) is used to generate summaries based on these descriptions, which aids in identifying implicit information from sequential text.
We use the embedding model (\emph{text-embedding-ada-002}) to obtain a $\lambda$-dimensional semantic representation for each node type that encode knowledge external to the domain by processing the summaries generated by the LLM.
OpenAI's \emph{text-embedding-ada-002} model is an LLM-based model specifically designed for embedding generation.
Finally, the embedding is fed into the MLP for dimensionality reduction:
\begin{equation}
	h^{LLM}_{t_n} = MLP(emb(P(t_n))), \forall t_n \in \mathbb{T},
\end{equation} 
where $emb$ is the text embedding model, $\mathbb{T}$ denotes the set of node types, and $h^{LLM}_{t_n} \in \mathbb{R}^\lambda$ indicates the latent representation vector of type $t_n$ enhanced by the external knowledge extracted through LLM.

Considering that the importance of different type-level information varies for fraudsters and benign entities, we design a strategy to adaptively learn representations based on the significance of type-level features.
This approach enables a more effective fusion of type-level representations into high-quality node representations, revealing the underlying dependencies between different types.
The weights to reflect the importance of each type-level representation are learned through a linear layer:
\begin{equation}
	\beta_{t_n}=\phi(W_t\cdot (h_{v_t}+h^{LLM}_{t_n})+b), 
\end{equation} 
where $W_t$ and $b$ learnable parameters, and $\phi$ is the sigmoid activation function.
Next, we generate the final representation by aggregating all type-level representations using a weighted average:
\begin{equation}
	z_{v_t}=\frac{1}{\mathbb{|T|}} (\sum_{t_n\in \mathbb{T}}^{} \beta_{t_n}\cdot h^{LLM}_{t_n}),
\end{equation} 
where $z_{v_t}$ denotes the final type-level representation for node $v_t$.

\subsection{Relation-level Enhancer}
To enhance the importance of fraudsters between different relations, we propose a relation-level enhancer based on LLMs.
Similarly to the type-level enhancer, we use LLMs to learn external knowledge from the different relations in graph fraud detection.
We equally construct a short relation-based description and an instruction that requires output in a fixed format, using them as prompts for the relations: $P(r_n)=$ \{Introduction to the relation $r_n$; Instruction\}.
Next, LLM (\emph{GPT-4}) generates summaries based on these descriptions and the embedding model constructs external semantic representations of the different relations.
The embedding is then fed into MLP for the purpose of dimensionality reduction:
\begin{equation}
	h^{LLM}_{r_n} = MLP(emb(P(r_n))), \forall r_n \in R,
\end{equation} 
where $emb$ is the text embedding model, $R$ denotes the set of relations, and $h^{LLM}_{r_n} \in \mathbb{R}^\gamma$ indicates the $\gamma$-dimensional latent representation vector of relation $r_n$ enhanced by the external knowledge extracted through LLM.

Since the contributions of different relations may vary significantly, we use an internal relation-attentive unit to adaptively assign weights to the representations of different relations for compression.
$\alpha_{r_n} = \sigma(W_r \cdot (h_{v_t}+h^{LLM}_{r_n}))$ is used as the learned attention coefficient, where $W_r$ is the trainable weight and $\sigma$ is \emph{LeakyReLU}.
For the target node $v_t$, we compute the overall node embedding $m_{v_t}$ of the relation-level enhancer as follows:
\begin{equation}
		m_{v_t}=\sum_{r_n \in R}^{} f_{att}(h^{LLM}_{r_n}, \delta_{r_n}  ), \quad \delta_{r_n}=\frac{exp(\alpha _{r_n}) }{ {\textstyle \sum_{r_n\in R}^{}} exp(\alpha _{r_n})},  
\end{equation} 
where $m_{v_t}$ indicates the final relation-level representation for node $v_t$, $\delta_{r_n}$ indicates the contribution of relation $r_n$ to the different embedding dimensions and $f_{att}$ represents the internal relation-attentive function.

\subsection{Fraud Detection}
To fuse multiple levels of representations, we use learnable weights to capture the importance of different levels and perform a weighted sum over the representations:
\begin{equation}
		F_{v_t}=h_{v_t}+w_{tp}\cdot z_{v_t}+w_{re}\cdot m_{v_t},  
\end{equation} 
where $F_{v_t}$ denotes the final node representation, $w_{tp}$ and $w_{re}$ are learnable weights.
Then, we feed it into the classifier of existing graph fraud detection methods to predict the label of the target node $v_t$:
\begin{equation}
	\hat{y}_t = \psi(W_{f}^TF_{v_t} + b_f),
\end{equation}
where $W_{f} $ is the trainable weight matrix, and $b_f$ is the bias vector. 
$\psi$ denotes the activation function \emph{softmax}. 
For all the graph fraud detection methods, we utilize the cross entropy loss to train the model:
\begin{equation}
    \mathcal{L}=-\sum_{t\in \mathcal{V}}y_t log\hat{y}_t+(1-y_t) log(1-\hat{y}_t).
\end{equation}

\subsection{Time Complexity Analysis}
We provide a theoretical analysis of the time complexity to demonstrate that the proposed MLED framework does not introduce additional time overhead, which can be detailed as follows.
The feature mapping layer involves a linear transformation with complexity $\mathcal{O}(N \cdot D \cdot U)$, where $N$ is the number of nodes, $D$ is the input feature dimension, and $U$ is the hidden dimension. 
For the type-level enhancer, the LLM embedding generation is treated as a constant $\mathcal{O}(1)$ per type due to the use of pre-trained models, while the MLP and aggregation operations contribute $\mathcal{O}(|\mathbb{T}| \cdot d_{in} \cdot d_{out} + N \cdot T)$, where $d_{in}$ and $d_{out}$ are the input and output dimensions of the MLP, and $T$ is the number of types. 
Similarly, the relation-level enhancer has a complexity of $\mathcal{O}(|R| \cdot d_{in} \cdot d_{out} + N \cdot R \cdot d)$, where $R$ is the number of relations and $d$ is the embedding dimension. 
Finally, the classification layer involves a linear transformation with complexity $\mathcal{O}(N \cdot U)$. 
The overall complexity of MLED is dominated by $\mathcal{O}(N \cdot (D \cdot U + T + R \cdot d))$, which is linear in the number of nodes $N$. 
Compared to benchmark graph fraud detection methods, MLED introduces negligible overhead as demonstrated in the efficiency analysis in Section 4.4, which typically has a complexity of $\mathcal{O}(N \cdot L \cdot d^2)$ for $L$-layer aggregation.



\section{Experiments}


\begin{table*}[]
\setlength\tabcolsep{7pt}
\caption{performance (\%) on two multi-relation datasets with three evaluation metrics.}
\begin{tabular}{lccccccc}
\toprule
\multirow{2}{*}{Methods} & \multicolumn{3}{c}{Amazon}                & \multicolumn{3}{c}{YelpChi}                 & \multirow{2}{*}{Avg. Impr ($\uparrow$)} \\ \cmidrule(lr){2-4} \cmidrule(lr){5-7}
                         & AUCROC       & AUCPRC       & F1-Macro     & AUCROC       & AUCPRC        & F1-Macro     &                           \\ \midrule
BWGNN                    & 88.56 ± 0.87 & 79.26 ± 1.11 & 90.48 ± 0.98 & 77.62 ± 2.37 & 39.87 ± 1.79  & 66.54 ± 0.73 & \multirow{2}{*}{+3.16\%}   \\
BWGNN+MLED               & 89.91 ± 0.95 & 81.05 ± 0.51 & 91.63 ± 0.38 & 79.92 ± 1.73 & 42.34 ± 1.21  & 69.71 ± 0.89 &                           \\ \midrule
GHRN                     & 88.35 ± 2.03 & 74.50 ± 4.57 & 86.35 ± 2.60 & 75.33 ± 1.44 & 34.53 ± 2.83  & 63.62 ± 1.41 & \multirow{2}{*}{+4.09\%}   \\
GHRN+MLED                & 89.77 ± 1.32 & 76.32 ± 1.61 & 87.93 ± 1.52 & 78.12 ± 0.81 & 37.79 ± 1.55  & 67.23 ± 0.45 &                           \\ \midrule
GDN                      & 92.16 ± 0.12 & 81.87 ± 0.17 & 89.75 ± 0.05 & 75.92 ± 0.51 & 38.04 ± 0.83  & 64.81 ± 0.25 & \multirow{2}{*}{+2.71\%}   \\
GDN+MLED                 & 93.68 ± 0.47 & 83.36 ± 0.73 & 91.11 ± 0.17 & 77.26 ± 1.12 & 40.31 ± 0.36  & 67.11 ± 0.92 &                           \\ \midrule
GAGA                     & 82.61 ± 2.87 & 56.69 ± 2.60 & 76.85 ± 1.08 & 71.61 ± 1.13 & 31.96 ± 3.37  & 61.81 ± 1.69 & \multirow{2}{*}{+2.93\%}   \\
GAGA+MLED                & 83.96 ± 1.73 & 58.42 ± 1.21 & 78.57 ± 0.76 & 73.22 ± 1.07 & 33.41 ± 0.78  & 64.22 ± 0.54 &                           \\ \midrule
DiG-In-GNN               & 86.30 ± 0.67 & 65.22 ± 1.09 & 82.61 ± 0.21 & 74.40 ± 0.93 & 37.06  ± 1.43 & 63.36 ± 0.27 & \multirow{2}{*}{+4.28\%}   \\
DiG-In-GNN+MLED          & 87.68 ± 0.92 & 67.43 ± 0.93 & 84.34 ± 0.47 & 77.22 ± 0.88 & 40.52  ± 1.21 & 66.82 ± 0.62 &                           \\ \midrule
ConsisGAD                & 93.91 ± 0.58 & 83.33 ± 0.34 & 90.03 ± 0.53 & 83.36 ± 0.53 & 47.33 ± 0.58  & 69.72 ± 0.30 & \multirow{2}{*}{+3.41\%}   \\
ConsisGAD+MLED           & 95.56 ± 0.30 & 84.79 ± 0.71 & 91.71 ± 0.39 & 85.02 ± 0.94 & 51.38 ± 1.45  & 72.88 ± 0.81 &                           \\ \bottomrule
\end{tabular}
\label{tab:results}
\end{table*}

\subsection{Experimental Setup}
\noindent{\bfseries Datasets.} We use the Amazon \cite{amazon}, YelpChi \cite{yelp}, T-Finance \cite{BWGNN}, and T-Social \cite{BWGNN} datasets to evaluate the performance of MLED. 
All the datasets consist of users and their associated comments. 
In the Amazon dataset, users are represented as nodes and there are three distinct relations between the nodes. 
The fraudsters are users with less than 20\% helpful votes.
To differentiate fraudsters from normal users, we label users with fewer than 20\% helpful votes as fraudsters.
The YelpChi dataset treats reviews as nodes, classifying them into two categories: spam and legit.
The connections between nodes are also of three types.
T-Finance aims to detect fraud accounts in transaction networks. 
The graph connections represent pairs of accounts with recorded transactions.
T-Social aims to identify fraud users in social networks. 
Two users are connected if they have been friends for over three months.

\noindent{\bfseries Baselines.} 
We select six state-of-the-art graph fraud detection baselines, including BWGNN \cite{BWGNN}, GHRN \cite{GHRN}, GDN \cite{GDN}, GAGA \cite{GAGA}, DiG-In-GNN \cite{DiG}, and ConsisGAD \cite{consisGAD}.
Then, we apply the proposed MLED framework to these baselines for extensive comparative experiments.
To ensure a fair comparison, we use the same settings in all baseline methods enhanced with MLED.
The details of the state-of-the-art methods are introduced in the supplementary material.
All the baselines are implemented using the source codes released by the authors. 
All experiments are conducted on Ubuntu 20.04.4 LTS with NVIDIA GTX 3090 Ti GPU and 128GB RAM. 
The MLED framework is implemented based on Python 3.8.20, Pytorch 2.4.0, and DGL 1.1.3.

\noindent{\bfseries Experimental Settings.}
In the experiments, we follow the hyperparameters given in the baseline papers with the Adam optimizer.
Besides, the sizes of the type-level representations and the relation-level representations in GPT\emph{-4} are controlled by two hyperparameters $\lambda$ and $\gamma$, which are set to sizes of $\lambda$=8 and $\gamma$=16, respectively.
To improve training efficiency, we employ mini-batch training with a batch size of 256 for Amazon, 1024 for YelpChi, 128 for T-Finance and T-Social.
For all baselines, we conduct a 10-run experiment to evaluate their performance, with the training ratio set to 1\%.

\noindent{\bfseries Evaluation Metrics.}
The overall performance is evaluated with three metrics: AUCROC, AUCPRC, and F1-Macro. 
AUCROC is calculated based on the relative ranking of prediction probabilities across all instances, which makes it robust to the label-imbalance problem.
AUCPRC evaluates the relationship between precision and recall across various thresholds.
F1-Macro represents the unweighted mean of the F1-scores for each label, where the F1-score is the harmonic mean of precision and recall.
Note that better performance is larger value of all metrics.

\begin{table}[]
\caption{Ablation study results on MLED with various LLM variants.}
\begin{tabular}{lccc}
\toprule
\multirow{2}{*}{Methods} & \multicolumn{3}{c}{Amazon}                                            \\ \cmidrule{2-4} 
                         & AUCROC                & AUCPRC                & F1-Macro              \\ \midrule
ConsisGAD                & 93.91 ± 0.58          & 83.33 ± 0.34          & 90.03 ± 0.53          \\
+ MLED\textsubscript{\textit{\footnotesize llama3-8B}}        & 94.53 ± 0.71          & 83.41 ± 0.39          & 90.89 ± 0.61          \\
+ MLED\textsubscript{\textit{\footnotesize GPT-3}}             & 94.34 ± 0.47          & 83.92 ± 0.43          & 91.11 ± 0.46          \\
+ MLED\textsubscript{\textit{\footnotesize GPT-3.5}}          & 94.73 ± 0.51          & 83.88 ± 0.67          & 91.03 ± 0.26          \\
+ MLED\textsubscript{\textit{\footnotesize GPT-4}}             & \textbf{95.56 ± 0.30} & \textbf{84.79 ± 0.71} & \textbf{91.71 ± 0.39} \\ \bottomrule
\end{tabular}

\begin{tabular}{lccc}
\toprule
\multirow{2}{*}{Methods} & \multicolumn{3}{c}{YelpChi}                                           \\ \cmidrule{2-4} 
                         & AUCROC                & AUCPRC                & F1-Macro              \\ \midrule
ConsisGAD                & 83.36 ± 0.53          & 47.33 ± 0.58          & 69.72 ± 0.30          \\
+ MLED\textsubscript{\textit{\footnotesize llama3-8B}}        & 84.15 ± 0.74          & 50.72 ± 1.41          & 71.94 ± 0.58          \\
+ MLED\textsubscript{\textit{\footnotesize GPT-3}}            & 84.79 ± 0.45          & 51.03 ± 1.63          & 71.73 ± 0.63          \\
+ MLED\textsubscript{\textit{\footnotesize GPT-3.5}}           & 84.67 ± 0.67          & 50.89 ± 1.21          & 72.13 ± 0.51          \\
+ MLED\textsubscript{\textit{\footnotesize GPT-4}}             & \textbf{85.02 ± 0.94} & \textbf{51.38 ± 1.45} & \textbf{72.88 ± 0.81} \\ \bottomrule
\end{tabular}
\label{tab:ablation}
\end{table}

\subsection{Overall Performance}
According to the experimental results of applying MLED to other methods, we analyze its overall performance. 
Table \ref{tab:results} reports the performance of the state-of-the-art methods with the application of our proposed MLED on Amazon and YelpChi datasets. 
In general, it can be observed that when applying our framework to existing methods, the performance is significantly improved, which proves the effectiveness of MLED.
Particularly, the latest state-of-the-art method ConsisGAD improves AUCROC, AUCPRC, and F1-Macro by 1.77\%, 1.76\%, and 2.11\% on the Amazon dataset after applying MLED, while that of the YelpChi dataset improves the AUCROC, AUCPRC, and F1-Macro by 1.99\%, 8.55\% and 4.53\% respectively.
Notably, the latest state-of-the-art method ConsisGAD improves AUCROC, AUCPRC, and F1-Macro by 1.77\%, 1.76\%, and 2.11\% on the Amazon dataset after applying MLED, respectively. 
Additionally, it increases AUCROC, AUCPRC, and F1-Macro by 1.99\%, 8.55\%, and 4.53\% on the YelpChi dataset, respectively.
Meanwhile, the results show that the six methods achieve average performance improvements of 3.16\%, 4.09\%, 2.71\%, 2.93\%, 2.93\%, 4.28\%, and 3.41\% across all evaluation metrics on the two datasets, respectively.
These results demonstrate the superiority and rationality of the proposed framework to integrate LLMs with graph structure information for fraud detection and enhance the ability to distinguish fraudsters.

\begin{table*}[]
\setlength\tabcolsep{7pt}
\caption{performance (\%) on two single-relation datasets with three evaluation metrics.}
\begin{tabular}{lccccccc}
\toprule
\multirow{2}{*}{Methods} & \multicolumn{3}{c}{T-Finance}                & \multicolumn{3}{c}{T-Social}                 & \multirow{2}{*}{Avg. Impr ($\uparrow$)} \\ \cmidrule(lr){2-4} \cmidrule(lr){5-7}
                         & AUCROC       & AUCPRC       & F1-Macro     & AUCROC       & AUCPRC        & F1-Macro     &                           \\ \midrule
BWGNN                    & 93.08 ± 1.57 & 77.79 ± 3.87 & 86.97 ± 1.51 & 84.40 ± 3.01 & 49.96 ± 3.75 & 76.37 ± 1.82 & \multirow{2}{*}{+1.97\%
}   \\
BWGNN+MLED               & 94.12 ± 0.78 & 78.98 ± 2.21 & 88.06 ± 1.43 & 85.89 ± 2.11 & 51.67 ± 1.63 & 78.45 ± 1.91 &                           \\ \midrule			
GHRN                     & 91.93 ± 0.93 & 65.94 ± 4.38 & 80.05 ± 4.43 & 84.20 ± 3.91 & 37.04 ± 11.02 & 71.25 ± 4.32 & \multirow{2}{*}{+2.23\%
}   \\
GHRN+MLED                & 93.37 ± 0.93 & 67.32 ± 2.11 & 81.49 ± 2.68 & 85.62 ± 2.46 & 38.73 ± 7.58 & 72.48 ± 3.46 &                           \\ \midrule
GDN                      & 88.75 ± 1.79 & 54.27 ± 4.31 & 76.62 ± 3.90 & 67.69 ± 1.49 & 47.39 ± 0.79 & 55.76 ± 0.82 & \multirow{2}{*}{+2.41\%
}   \\
GDN+MLED                 & 89.93 ± 1.26 & 56.63 ± 2.36 & 77.83 ± 1.73 & 68.95 ± 1.26 & 48.71 ± 0.68 & 57.21 ± 0.44 &                           \\ \midrule
GAGA                     & 92.36 ± 1.45 & 64.34 ± 6.01 & 81.10 ± 2.60 & 78.92 ± 1.26 & 23.72 ± 4.81 & 65.58 ± 3.30 & \multirow{2}{*}{+2.68\%
}   \\	
GAGA+MLED                & 93.87 ± 1.71 & 66.21 ± 3.37 & 82.63 ± 1.79 & 80.32 ± 1.34 & 25.11 ± 3.43 & 66.89 ± 2.61 &                           \\ \midrule
DiG-In-GNN               & 92.53 ± 0.31 & 79.50  ± 0.35 & 88.51 ± 0.12 & 93.75 ± 0.45 & 68.73  ± 1.16 & 81.40 ± 0.78 & \multirow{2}{*}{+1.63\%
}   \\
DiG-In-GNN+MLED          & 93.78 ± 0.48 & 80.78  ± 0.71 & 89.96 ± 0.36 & 94.96 ± 0.61 & 70.22 ± 1.37 & 82.79 ± 0.63 &                           \\ \midrule
ConsisGAD                & 95.33 ± 0.30 & 86.63 ± 0.44 & 90.97 ± 0.63 & 94.31 ± 0.20 & 58.38 ± 2.10 & 78.08 ± 0.54 & \multirow{2}{*}{+1.87\%
}   \\
ConsisGAD+MLED           & 96.97 ± 0.25 & 87.85 ± 0.32 & 92.21 ± 0.37 & 95.79 ± 0.26 & 60.21 ± 1.73 & 79.66 ± 0.34 &                           \\ \bottomrule
\end{tabular}
\label{tab:results_single}
\end{table*}

To verify the generalization capability and robustness of the MLED framework on single-relation graph data, we conduct experiments on two benchmark single-relation fraud detection datasets, as shown in Table \ref{tab:results_single}.
The results show that MLED still significantly improves the performance of the baseline methods on the single-relation datasets.
The method ConsisGAD improves AUCROC, AUCPRC, and F1-Macro by 1.72\%, 1.41\%, and 1.37\% on the T-Finance dataset, and 1.57\%, 3.13\%, and 2.02\% on the T-Social dataset, respectively, after applying MLED. 
The results show that even if the single-relation graphs lack complex topology, MLED is still able to compensate for the lack of structural information in the single-relation graphs through the external knowledge extracted by LLM, which improves the fraud detection performance.
Overall, MLED performs well on both multi-relation and single-relation data, which proves its generalizability and robustness as a general framework.

\subsection{Ablation Analysis}
\begin{figure}[th]
	\centering
    \includegraphics[width=\linewidth]{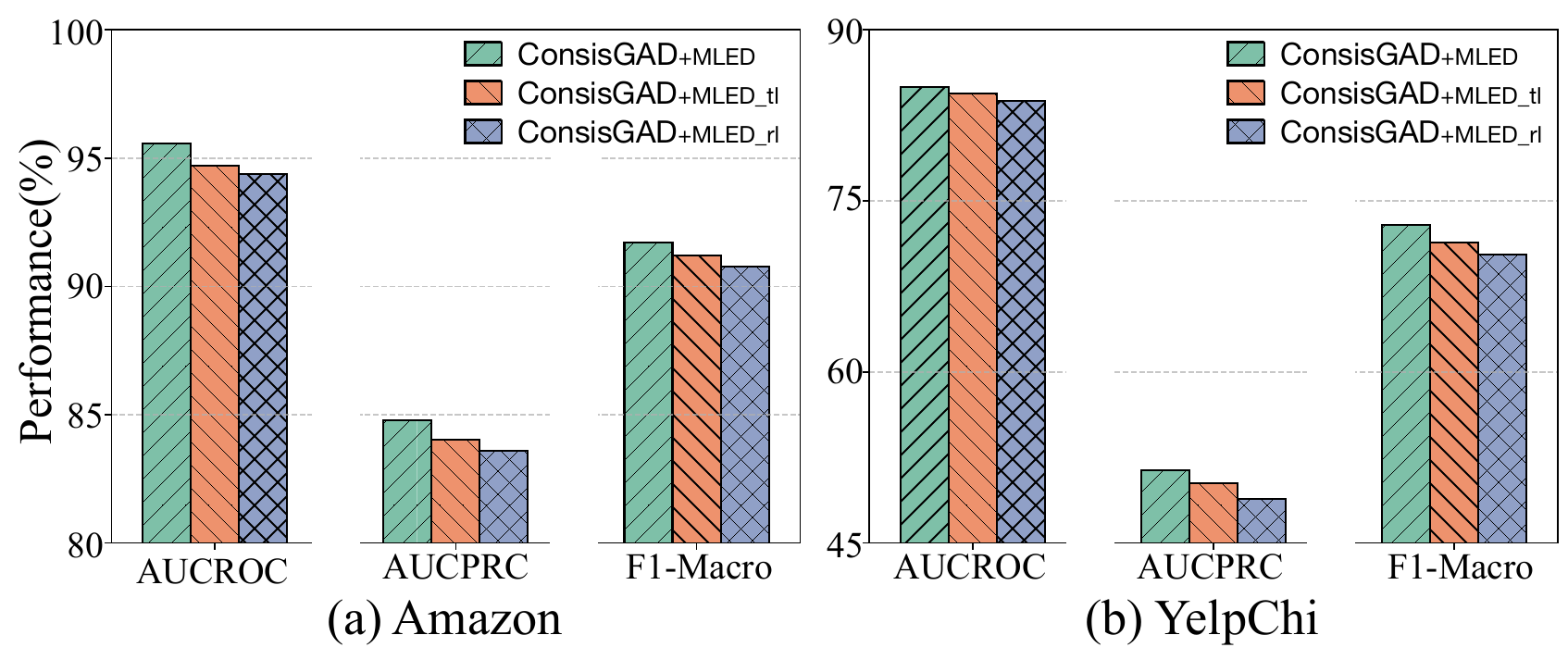}
	\caption{Ablation study results on various MLED variants with the SOTA method.}
	\label{ablation}
\end{figure}

To investigate how the two level enhancers improve the performance of the proposed framework, we conduct an ablation analysis on the following variants of MLED. 
MLED\_tl and MLED\_rl denote the models that contain only the type-level enhancer and the relation-level enhancer, respectively.
Figure \ref{ablation} illustrates the experimental results of ConsisGAD in applying MLED and its variants.
The results represent that the multi-level structure of framework outperforms both one-level variants in all metrics.
The three evaluation metrics decreased by 0.9\%, 0.92\% and 0.55\% on the Amazon dataset with only type-level enhancer, which decreased by 1.27\%, 1.41\% and 1.05\% with only relation-level enhancer.
Similarly, on the YelpChi dataset, the three evaluation metrics decreased by 0.73\%, 2.33\% and 2.17\% with only type-level enhancer, while those decreased by 1.48\%, 5.22\% and 3.71\% with only relation-level enhancer.
It is noticeable that the type-level and relation-level enhancers play critical roles in MLED and their combination greatly benefits the fraud detection. 
It also demonstrates the effectiveness of the two proposed LLM enhancers in MLED.
Moreover, the representation obtained from the type-level enhancer makes a greater contribution to our model.
The results indicate that continuously widening the gap between fraudsters and benign entities is the key to detecting fraudsters.

\begin{figure}[th]
	\centering
    \includegraphics[width=\linewidth]{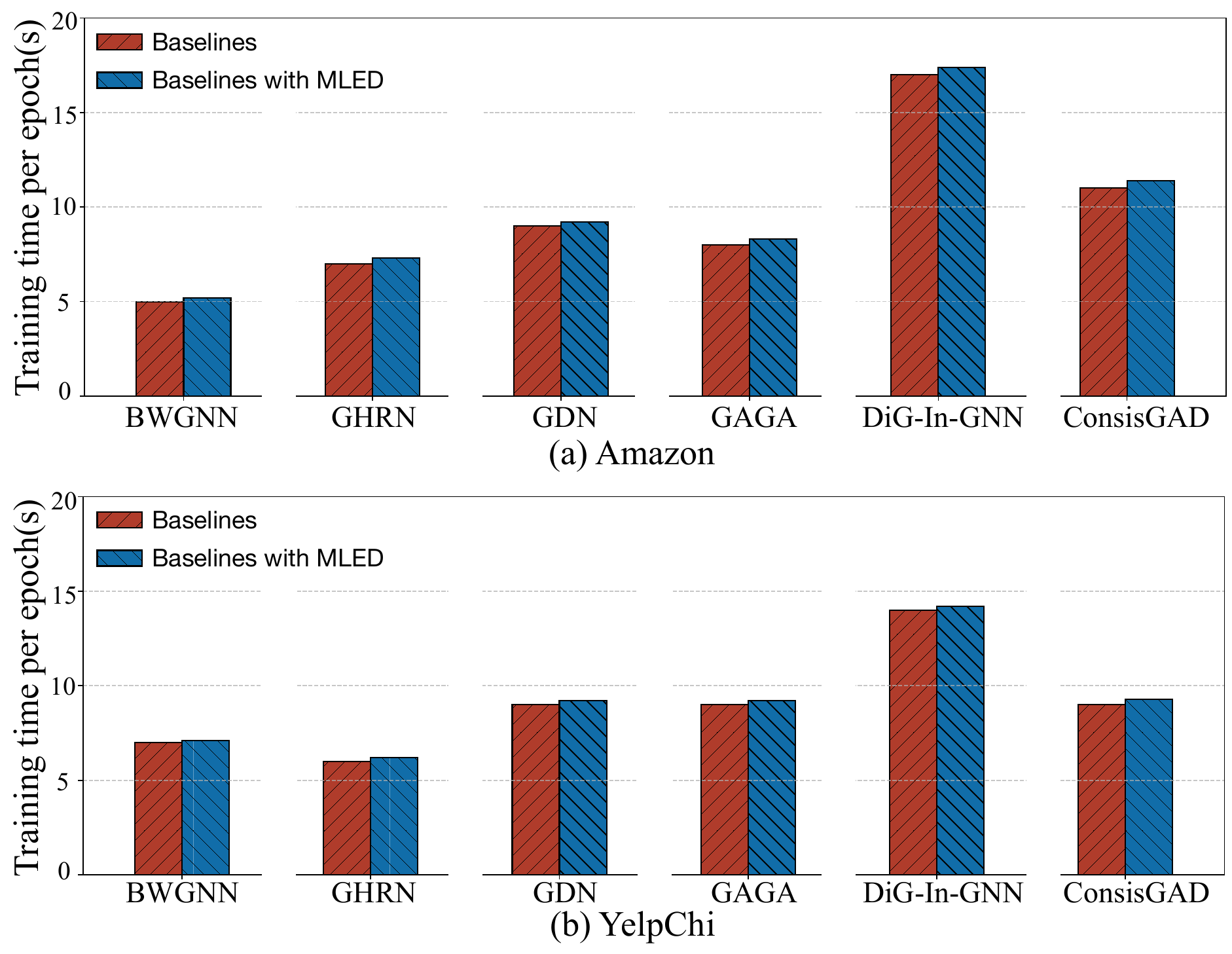}
	\caption{Efficiency analysis on the Amazon and YelpChi datasets.}
	\label{training_time}
\end{figure}

We conduct additional ablation experiments to assess the impact of different LLM variants on the MLED and validate whether the scale and capability of LLMs significantly affect semantic comprehension.
The experiments investigate four LLM variants within the MLED framework and apply them to the latest state-of-the-art method, which maintains the same graphical encoder configuration, as shown in Table \ref{tab:ablation}.
In general, all LLM variants improve the performance of the baseline method, but the extent of improvement varies between models.
\emph{GPT-4} achieves the highest performance gains on both datasets, while smaller models such as \emph{Llama3-8B} and \emph{GPT-3.5} exhibit modest improvements but still underperform compared to \emph{GPT-4}.
The results highlight the critical importance of LLM scale and reasoning capability in  MLED.
The proposed MLED framework has superior ability to detect fraudsters when it employs LLM with more advanced semantic understanding and broader knowledge coverage.

\subsection{Efficiency Analysis}
To further compare the efficiency of different methods when applying MLED, we calculate the average training time during training on the Amazon dataset.
Figure \ref{training_time} presents a comparison between the average training time per epoch of baselines and their corresponding versions enhanced by MLED.
On the Amazon dataset, the six baselines increase in time by only 0.17\emph{s}, 0.21\emph{s}, 0.37\emph{s}, 0.31\emph{s}, 0.38\emph{s}, and 0.26\emph{s} after applying MLED, while they increase in time by only 0.1\emph{s}, 0.18\emph{s}, 0.17\emph{s}, 0.21\emph{s}, 0.19\emph{s}, and 0.21\emph{s} on the YelpChi dataset.
The results show that all methods introduce only negligible time overhead after applying MLED.
This demonstrates that our framework exhibits excellent efficiency and scalability, which is mainly due to the simplicity of our multi-level enhancers.

\begin{figure*}[th]
	\centering
    \includegraphics[width=\linewidth]{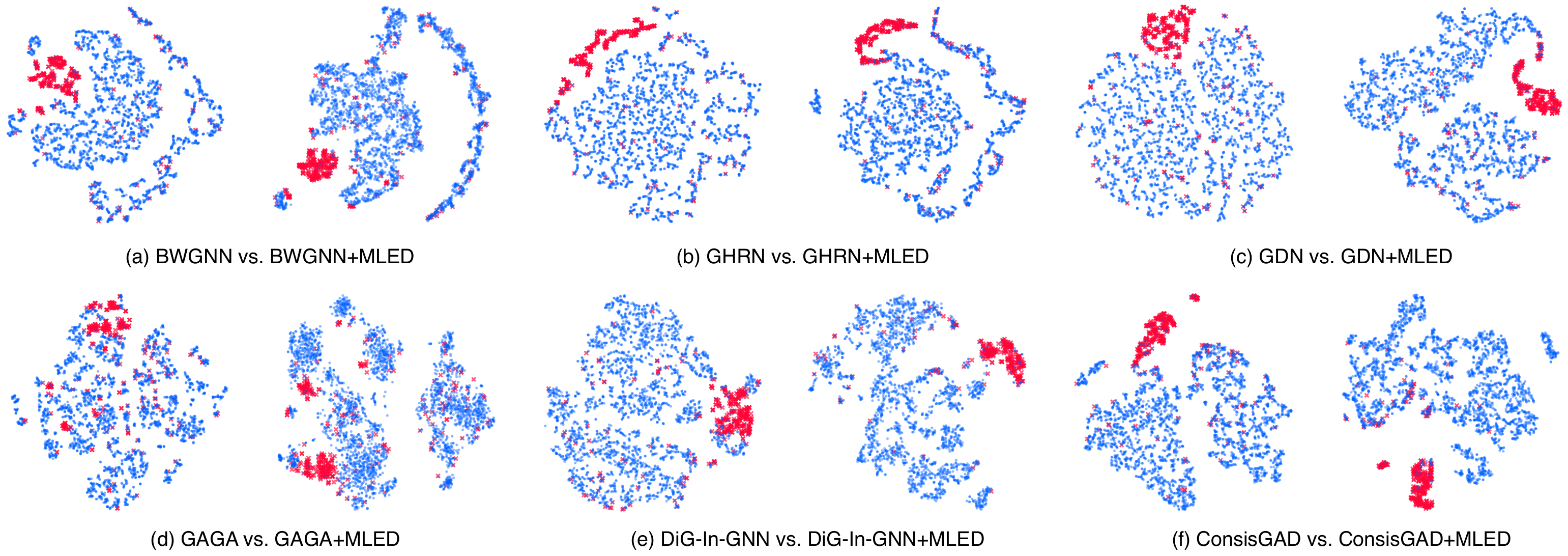}
	\caption{The visualization analysis on the Amazon dataset, where red represents fraudsters blue represents benign entities. Each pair of subfigures consists of baselines (left) and baselines enhanced by MLED (right).}
	\label{visualization}
\end{figure*}

\subsection{Visualization Analysis}
To demonstrate the effectiveness of the proposed framework more directly, we conduct a visualization task with Amazon dataset.
We present the visualization results of baselines and their enhanced versions in six subfigures for comparison, as shown in Figure \ref{visualization}.
Specifically, MLED makes fraudsters form compact clusters that are more distinctly separated from benign entities.
Figures \ref{visualization} (d) and Figures \ref{visualization} (e) show the most obvious comparison results.
In the original method, fraudsters form loose clusters and blend with benign entities, whereas they exhibit tight clusters with significantly higher differentiation after MLED enhancement.
The visualizations empirically demonstrate that MLED enhances the discriminative capacity of graph fraud detection methods by leveraging multi-level LLM-driven knowledge.
The clearer fraud separation further validates the effectiveness of the proposed framework in real-world scenarios.

\subsection{Parameter Analysis}
\begin{figure}[th]
	\centering
    \includegraphics[width=\linewidth]{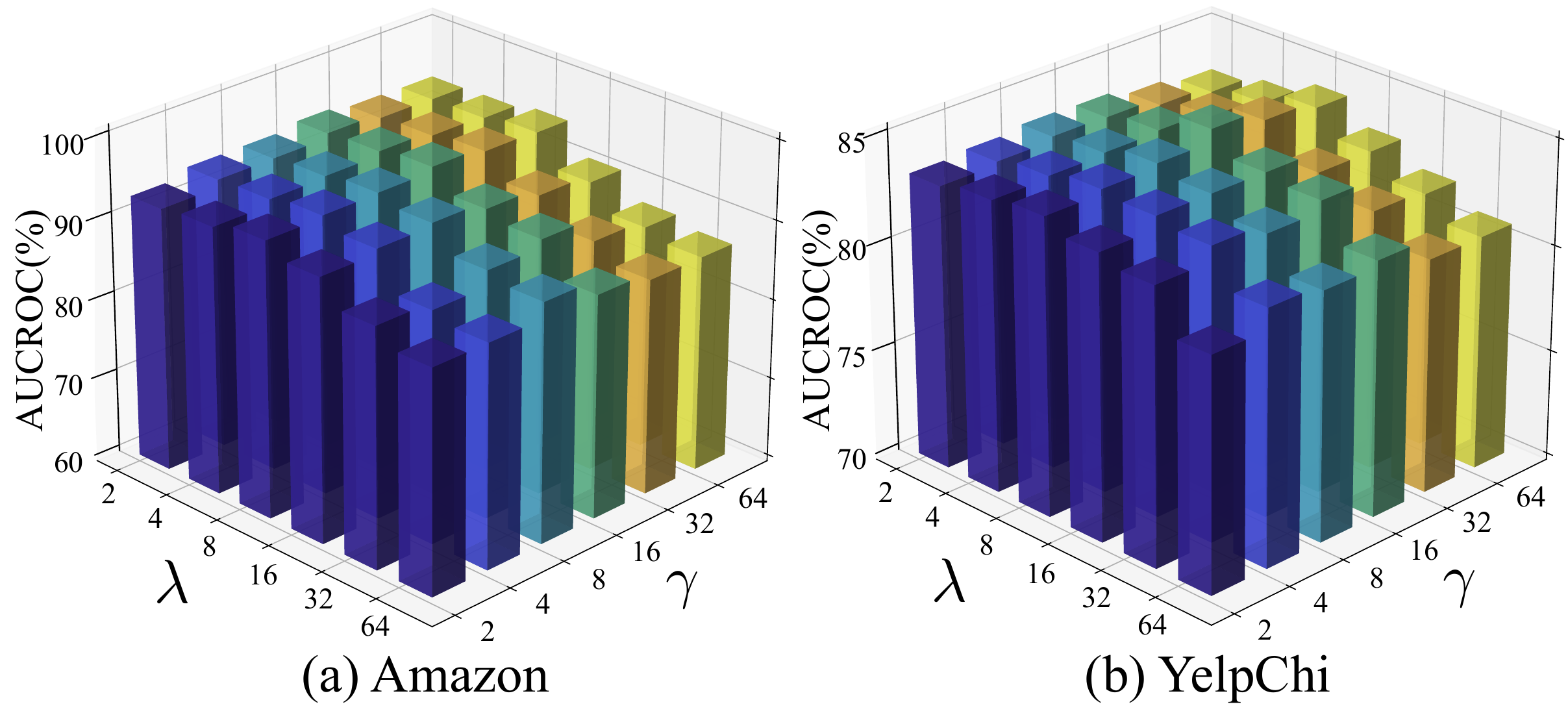}
	\caption{Parameter analysis of $\lambda$ and $\gamma$ on Amazon and YelpChi datasets. Note that the values of $\lambda$ and $\gamma$ range from
[2, $\ldots$, 64].}
	\label{sensitivity}
\end{figure}

The proposed MLED framework contains two main hyperparameters, i.e., $\lambda$ and $\gamma$, which indicate the size of type-level representations and relation-level representations, respectively.
To evaluate the robustness of MLED under varying hyperparameter configurations, we conduct a parameter sensitivity analysis on the Amazon and YelpChi datasets. 
This experiment aims to analyze the impact of the two hyper-parameters on fraud detection performance.
Specifically, we vary the values of $\lambda$ and $\gamma$ within the range of
[2, $\ldots$,64] and Figure \ref{sensitivity} presents the experimental results.
Based on the observations in the figure, we make several conclusions.
Firstly, MLED tends to exhibit suboptimal performance when $\lambda$ and $\gamma$ are set to low values, e.g., 2 and 4.
This emphasizes the crucial role of both representations in the MLED framework and highlights their effectiveness.
Secondly, we can observe that excessively high values of $\lambda$ and $\gamma$ can also adversely affect performance, because they may mask the importance of the original node representations.
Finally, MLED demonstrates relatively stable AUCROC performance across a wide range of $\lambda$ and $\gamma$ values, which proves its robustness.

\section{Conclusions}
In this paper, we propose a novel multi-level LLM enhanced framework called MLED for graph fraud detection, which is applicable to existing methods as a general framework.
Accordingly, we utilize the power of LLMs to extract external knowledge from textual information to enhance type-level and relation-level representations.
The framework’s fusion of multi-level representations effectively enhances the distinction between fraudsters and benign entities, as well as the importance of different relations to fraudsters.
The experimental results demonstrate that MLED enhances the performance of the graph fraud detection model, which validates the effectiveness of the proposed framework.

\begin{acks}
The work is supported by the National Natural Science Foundation of China (No. 62272487 and 62433016).
\end{acks}

\bibliographystyle{ACM-Reference-Format}
\bibliography{camera.bib}


\end{document}